# Depth Estimation Analysis of Orthogonally Divergent Fisheye Cameras with Distortion-Removal

Matvei Panteleev, Houari Bettahar, Member, IEEE

Department of Electrical Engineering and Automation, School of Electrical Engineering, Aalto University, Espoo, Finland.

Corresponding author: Houari Bettahar (e-mail: houari.bettahar@aalto.fi).

**ABSTRACT** Stereo vision systems have become popular in computer vision applications, such as 3D reconstruction, object tracking, and autonomous navigation. However, traditional stereo vision systems that use rectilinear lenses may not be suitable for certain scenarios due to their limited field of view. This has led to the popularity of vision systems based on one or multiple fisheye cameras in different orientations, which can provide a field of view of 180x180 degrees or more. However, fisheye cameras introduce significant distortion at the edges that affects the accuracy of stereo matching and depth estimation. To overcome these limitations, this paper proposes a method for distortion-removal and depth estimation analysis for stereovision system using orthogonally divergent fisheye cameras (ODFC). The proposed method uses two virtual pinhole cameras (VPC), each VPC captures a small portion of the original view and presents it without any lens distortions, emulating the behavior of a pinhole camera. By carefully selecting the captured regions, it is possible to create a stereo pair using two VPCs. The performance of the proposed method is evaluated in both simulation using virtual environment and experiments using real cameras and their results compared to stereo cameras with parallel optical axes. The results demonstrate the effectiveness of the proposed method in terms of distortion removal and depth estimation accuracy.

**INDEX TERMS** Divergent stereo vision, Fisheye camera, Distortion removal, Camera models,

## I. INTRODUCTION

Camera arrays systems have been extensively used in various fields, such as robotics, autonomous vehicles, and medical imaging, for depth estimation and 3D reconstruction. For instance, they have been utilized in NASA's rovers for navigation purposes for a considerable period [1], [2]. A conventional rover overview system comprises numerous pairs of cameras that aid in navigation and environmental evaluation. These camera pairs work together to create a depth map that corresponds to the distance and space around the rover. The Yandex autonomous delivery robot developed by Yandex is equipped with 4 ultra-wide-angle cameras located on its front, back, and sides, which offer an extensive panoramic view for the operator and the odometry algorithms[3]. Similarly, various car manufacturers incorporate camera-based surround-view systems to facilitate parking or autonomous driving, with camera positions varying between models and brands. Nonetheless, the general approach is to position cameras to maximize the coverage of the surrounding area. This has led to the popularity of fisheye lenses, which can provide a field of view of 180x180 degrees or more. However, fisheye lenses introduce significant distortion that affects the accuracy of stereo matching and depth estimation, unlike stereovision systems that rely on two images of the same scene, captured by cameras with parallel optical axes.

Recently, several methods have been proposed to overcome the distortion issues associated with fisheye lenses. One of the most common methods was proposed for stereo fisheye camera systems aimed in one direction with parallel optical axes [4], [5]. Deep learning techniques have also been used to improve the accuracy and efficiency of stereo vision systems. For instance, convolutional neural networks (CNNs) have been applied to extract depth information from a pair of images [6]. Nevertheless, this method requires significant computational resources. Another method involves mounting two 245° cameras at opposite ends of a rigid rod and aiming them towards each other to achieve a circular depth perception volume with a 65° vertical field-of-view [7]. While this method produces a panoramic depth view with sufficient quality for autonomous navigation and UAV localization. However, it is difficult to implement for other types of robots [8] . The third method is the least explored so far and involves a "Special Stereo Vision System" developed by Zhang et al [9]. This system uses a system of four fisheye lenses cameras with more than 180° field of view placed at 90° angle to each other. The authors discussed calibration and the epipolar rectification method. However, they have not studied the depth estimation nor analyzed the accuracy of their method.

Despite the advancements in stereo vision techniques, there is still a need for robust and accurate stereo vision systems using divergent fisheye lenses cameras. This paper proposes a method for distortion-removal and depth estimation analysis for stereovision system using orthogonally divergent fisheye cameras (ODFC). The proposed method uses two virtual pinhole cameras (VPC), each VPC captures a small portion of the original view and presents it without any lens distortions, emulating the behavior of a pinhole camera. The proposed method was validated in both virtual environment and real environment using real cameras. The results demonstrate the effectiveness of the proposed method in terms of distortion removal and depth estimation accuracy.

The remainder of the paper is organized as follows: Section II models the Fisheye camera, Section III describes the distortion removal concept, Section IV presents both simulation and experimental results associated with depth analysis. Section V concludes the paper and discusses future work.

## II. FISHEYE CAMERA MODELS

The difficulties encountered when using existing stereo vision algorithms in ultrawide-angle cameras are due to the nature of their optical systems. These cameras have in their basis a complex system of lenses. The features of this system make it possible to achieve very high viewing angles, but also cause aberrations and signature image distortions. To describe the projection properties of a wide range of such cameras, researchers have resorted to approximations called camera models.

The camera model for a camera is a function that describes the transformations between points in three-dimensional space in the camera's reference frame ($P = [x_c \quad y_c \quad z_c]^T$) and points on the image plane ($p = [u \quad v]^T$). Most consumer cameras may be modelled with perspective projection, as in Fig.1, but wide field-of-view cameras require a more complicated description owing to the strong radial distortion.

As the distortions are considered radially symmetrical, most models work in the domain of distance from pixel to the center of an image $\rho$ and the angle of incidence of a point in 3D space $\theta$. For the implemented system only front projection function of a fisheye model, commonly denoted as $\pi_f(\cdot)$, is required.

### A. KANNALA-BRANDT MODEL

Kannala-Brandt's model [10] for lenses with radially symmetric distortion is implemented in OpenCV and many less popular libraries. The authors found that five terms of an odd power polynomial were sufficient to describe typical distortions as a relation between the angle of incidence and distance between projected pixel and the center of the image. Thus, radial relations of this model can be written with the following equation.

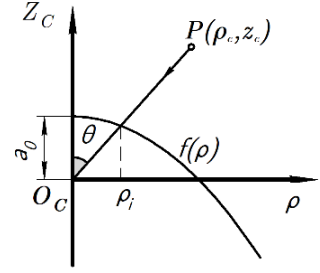

**Figure 1** : A 3D point in camera frame expressed in $z_c$ and $\rho$ coordinates. $f(\rho)$ represents a projection function in case of polynomial models, $a_0$ is a focus distance.

$$\rho = k_1\theta + k_2\theta^3 + k_3\theta^5 + k_4\theta^7 + k_5\theta^9, \quad (2)$$

where $k_1 \ldots k_5$ – model parameters.

This model can be calibrated using OpenCV and CamOdoCal libraries.

### B. MEI MODEL

The Mei model [11] is a more general version of the Geyer model [12]. It was originally designed to simulate catadioptric cameras more effectively, as it allows the use of different distortion functions to simulate mirrors of different shapes but has also proved to be highly suitable for wide-angle cameras. The model consists of projection rule (3) and distortion terms (4).

$$m_u = \begin{bmatrix} X_u \\ Y_u \end{bmatrix} = \begin{bmatrix} \dfrac{X_s}{Z_s + \xi} \\ \dfrac{Y_s}{Z_s + \xi} \end{bmatrix}, \quad (3)$$

where $\xi$ – model parameter, $X_s, Y_s, Z_s$ – normalized point coordinates.

The distortion terms with $k_1, k_2, p_1, p_2$ as model parameters:

$$m_d = m_u(k_1\rho + k_2\rho^2) + \begin{bmatrix} 2p_1 X_u Y_u + p_2(\rho + 2X_u^2) \\ 2p_2 X_u Y_u + p_1(\rho + 2Y_u^2) \end{bmatrix}. \quad (4)$$

Finally, the resulting pixel coordinates can be obtained as the sum of the terms:

$$\begin{bmatrix} u \\ v \end{bmatrix} = m_u + m_d. \quad (5)$$

This model may be calibrated using the CamOdoCal library.

### C. SCARAMUZZA MODEL

The Scaramuzza model [13], which is the basis of the Matlab Omnidirectional Camera Calibration Toolbox, is also widely used. It doesn't have a closed form back projection ruleIt associates points in the image with their corresponding points in the camera coordinates, as follows:

$$\begin{bmatrix} x_c \\ y_c \\ z_c \end{bmatrix} = \lambda \begin{bmatrix} u \\ v \\ a_0 + a_2\rho^2 + a_3\rho^3 + a_4\rho^4 \end{bmatrix}, \quad (6)$$

where $a_0 \ldots a_4$ – model parameters, $\lambda = \rho_c/\rho_i$ is a scaling factor, proportional to the ratio of the distance from the point to the optical axis $\rho_c$ and the distance from its pixel-projection to the center of the image $\rho_i$. As $\rho_i$ depends on yet unknown pixel coordinates, some approximation algorithm (e.g. Newton's method) has to be run on every point to obtain it and calculate the projection

$$\begin{bmatrix} u \\ v \end{bmatrix} = \frac{1}{\lambda} \begin{bmatrix} x_c \\ y_c \end{bmatrix}. \quad (7)$$

### D. ATAN MODEL

This model describes an ideal equidistant fisheye projection. It lacks the flexibility of other models because it has only one parameter - the field of view but serves as a good reference model as it is used in the simulated fisheye camera chosen for the experiments. The projection is expressed as follows:

$$\begin{bmatrix} u \\ v \end{bmatrix} = \begin{bmatrix} \dfrac{f_x \theta}{\sqrt{y_c^2/x_c^2 + 1}} \\ \dfrac{f_y \theta}{\sqrt{x_c^2/y_c^2 + 1}} \end{bmatrix}. \quad (8)$$

### III. Distortion-removal concept

The most popular and well-performing [14] stereovision algorithms are based on two images of the same scene, captured by cameras with parallel optical axes. To achieve this with divergent fisheye images, a virtual pinhole camera (VPC) needs to be introduced. A VPC captures only a small region of the original view and presents it as a pinhole camera would - with no lens distortions. By correctly selecting the regions in the two images, it is possible to form a stereo pair compatible with existing stereo matching algorithms using two VPCs (Fig. 2).

The process starts with identifying the desired VPC intrinsic parameters such as resolution and field of view. Then a blank image of the chosen resolution is generated and projected using the pinhole camera back projection function $\pi_p^{-1}(\cdot)$ into a 3D space as described in

$$[x_c \quad y_c \quad z_c]^T = [u \cdot z_c/f_x \quad v \cdot z_c/f_y \quad z_c]^T, (9)$$

where $x_c, y_c, z_c$ – are coordinates in the camera frame, $f_x, f_y$ are focus distances.

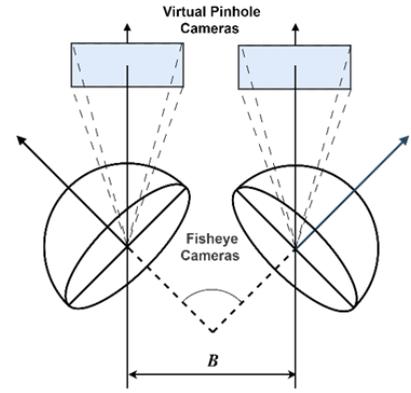

**Figure 2**: Principle of stereopair forming.

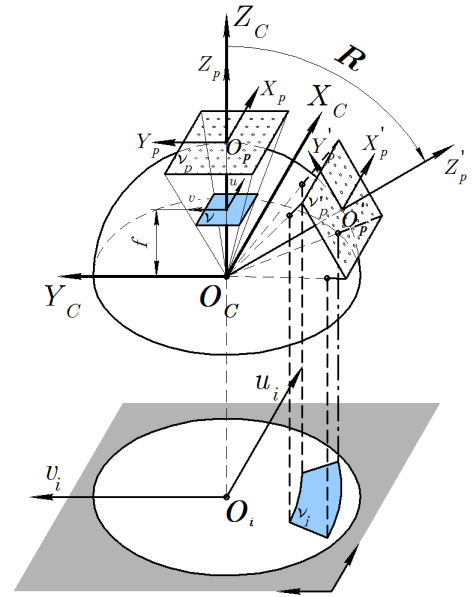

**Figure 3**: Geometric principle of the non-distortion process. $O_C X_C Y_C Z_C$ is the fisheye camera reference frame; $O_p X_p Y_p Z_p$ is the local frame of a pointcloud obtained pinhole back projection $v_p$; $R$ is the rotation of VPC relative to fisheye camera; $O'_p X'_p Y'_p Z'_p$ is the local frame of rotated pointcloud $v'_p$; $v_i$ is the patch of a fisheye image $u_i v_i$ containing pixels for distortion removal.

The resulting point cloud is then rotated towards the desired direction using a rotation matrix $R$. By using a fisheye camera model projection function $\pi_f(\cdot)$ to project the rotated point cloud, the resulting distorted image can be matched with every pixel in the original image. This process is depicted in Fig. 3. The resulting transformation from a VPC image pixel $p_p$ to the corresponding fisheye image pixel $p_i$ is

$$p_i = \pi_f\left(R \cdot \pi_p^{-1}(p_p)\right), \quad (10)$$

Running the process for every pixel in an image might be a computationally heavy process depending on the used model. To accelerate the initial projection, the process was parallelized and a lookup table was employed to store the computed results for each pixel in a memory structure.

These stored values could be reused for subsequent frames, eliminating the need for repeated calculations and enabling real-time performance.

## IV. Results and discussions

### A. Depth and depth error estimation

Using the abovementioned models presented in section II with the proposed concept for distortion removal on fisheye images, it is possible to get a pair of images for stereo matching. The output of a stereo matching is a depth map of the perceived space. The accuracy of stereo cameras can be determined by how closely this depth map conveys real-world information about the distance to surfaces.

From the resulting depth maps, 3D reconstruction of the scene can be performed. The result is represented as a point cloud, similar to that shown in Fig.4. For the purpose of accuracy estimation, the points outside the vicinity of the target plane are discarded. Next, 1000 random points are selected from the remaining points, normalizing the sample size for all distances. The remaining points are used to calculate the error. It is done by measuring the standard deviation and variance of the depth points from their assumed positions.

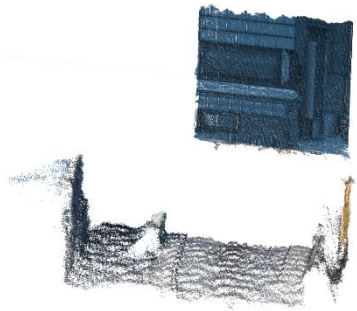

**Figure 4:** Example of a raw point cloud

### B. Simulation results

Reliably assessing accuracy can be problematic due to the influence of various factors. The primary sources of error in stereo cameras, as stated in [14], include sensor errors, measurement conditions, and properties of the observed surfaces. Sensor errors can result from inaccurate camera parameter selection and imperfections in optics, leading to incorrect distortion correction and systematic error in evaluating point coordinates in space. However, better optics and more precise calibration can minimize these effects. Measurement conditions, such as illumination, distance to the observed surface, and camera position, can also impact accuracy. Occlusion can occur in certain camera and object configurations, hindering depth computation, and increasing the distance between the stereo pair and the object can increase error. Lastly, the observed surfaces should have a pronounced texture and be lambert [15], [16].

To mitigate most of these problems, initial tests were conducted in the virtual environment of the Unity game engine. As a result, it is possible to place virtual cameras and targets precisely, conduct repeatable experiments, and control the environment. The only remaining variables in this setting are the distortion models and the distance to a target.

The virtual nature of the experiment makes it possible to precisely know the position of the target object and accurately determine the error. The surface properties are controlled by using different textures on the target object and averaging the results over them.

Since the greatest distortion in a fisheye image occurs at the edges of the image, and it's the setting in which the fields of view intersect most frequently, experimental stereo pair were constructed such that the angle between the main axes of the cameras is 90°.

Beside other advantages, the virtual environment enables to put several objects at the same coordinates. Thus, a pair of reference cameras with no distortion can be precisely aligned with VPCs and set to use the same parameters (see Fig. 5). Their accuracy will serve as reference for comparison (it is denoted as Reference).

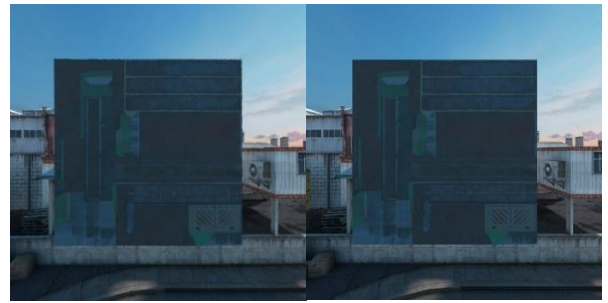

**Figure 5:** Image comparison. Image with removed distortions (left) and a shot from a reference camera (right). Only under close inspection, only small differences can be observed.

In the virtual fisheye camera setup 'lenses' had the same parameters, a field of view of 180°, and a baseline of 20 cm. All the presented models have been calibrated on one set of calibration pattern images captured in the virtual environment. The calibration results are in Table I. As the virtual cameras are identical, only one set of parameters is required.

TABLE I: Calibration results for Mei, Kannala-Brandt, and Scaramuzza and Atan models

| Model | Software | Mean Reprojection Error | Parameters |
|---|---|---|---|
| Mei | CamOdoCal | 0.102 | $\xi=1.474$, $p_1=1.66*10e-4$, $p_2=8*10e-5$, $k_1=-0.208, k_2=0.153$ |
| Kannala-Brandt | CamOdoCal | 0.099 | $k_2=7.58*10e-4$, $k_3=-3.26*10e-4$, $k_4=4.03*10e-5$, $k_5=-1.86*10e-6$, |
| Scaramuzza | MATLAB | 0.121 | $a_0=345.1$, $a_1=-0.0011$, $a_2=5.762*10-7$, $a_3=-1.398*10-9$ |
| Atan | - | - | Fov=180o |

One way to evaluate the effectiveness of distortion removal was to perform image subtraction. This method involved taking the difference between a reference image (without distortion) and the processed image with removed distortion, resulting in a mask that revealed potential defects, as illustrated in Fig. 6. The mean pixel intensity can be used to quantify this difference.

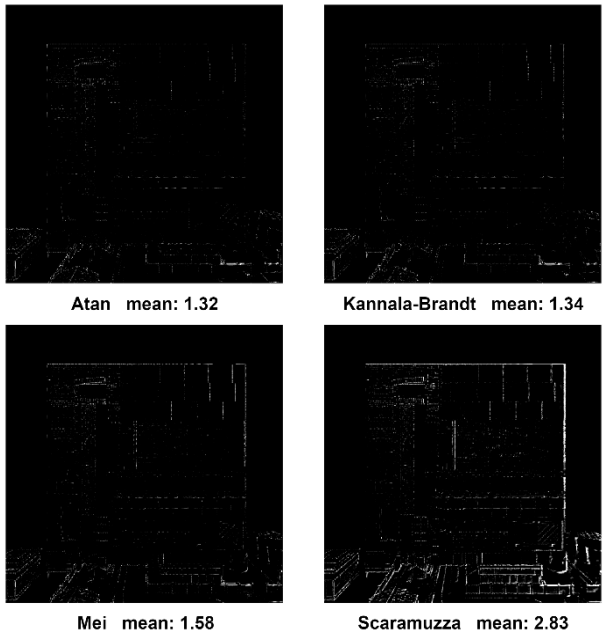

**Figure 6:** The difference in images increased the contrast and brightness for visibility. All images demonstrate only small defective areas concentrated around sharp edges. Scaramuzza demonstrates more bright areas than the other models, which can be explained by worse calibration quality.

A depth-quality evaluation was performed, and the results are displayed in Fig. 7. The x-axis shows the depth to the plane in terms of the number of stereopair baseline lengths, and the y-axis shows the root-mean-square (RMS) error between the estimated depth and the true depth, known from the target placement in the simulated world. We examined how the RMS error of the depth estimation varied with the depth for different distortion models: Scaramuzza, Mei, and Kannala-Brandt, as well as the case without distortion. We used a second order polynomial regression model to fit the data. The vertical lines show the standard deviation of the depth distribution in the pointcloud. Each line represents a different model and is calculated from the average of all the textures.

As can be seen from the figure, the best depth estimation result is naturally shown by the reference stereo pair. This is closely followed by the ideal model embedded in the virtual camera, which shows that even in the case of perfect correspondence between forward and backward projections, some information is lost or degraded. Of the models examined, the Kannala-Brandt model demonstrates the smallest error, almost identical to system with no distortion.

Mei model performs substantially worse, while Scaramuzza model demonstrates the worse performance overall. These experiments prove that the proposed stereo vision system design is feasible, and the decrease in quality is not substantial. The best results are shown by the Mei and Kannala-Brandt models.

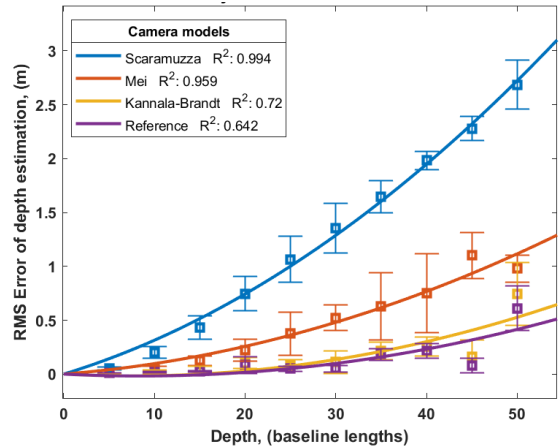

**Figure 7:** Depth quality evaluation of Mei, Kannala-Brandt and Scaramuzza models as well as, the reference virtual stereo cameras without distortion.

### C. Experimental results

After the virtual experiments the next step is the testing of the system with real cameras. For this purpose, 2 off-the-shelf cameras were used with the following parameters: 1.45 mm F2.2 1/1.8 FOV 190◦ (AC123B0145IRM12MM) mounted at an angle of 90◦ in a 3D printed body. The base of the stereo pair is approximately 72 mm. An image of the stereo-vision system module is shown in Fig. 8. An example of a fisheye image produced by such a camera module is shown in Fig. 9.

In the experiments the measurement condition error is minimized using artificial light, and the distance to the target object is controlled by markings on the experimental table. Similar to the virtual tests, three textures printed on a matte paper sheet were used to reduce the observed surface properties influence. Each camera has been individually calibrated against checkerboard pattern images, and the resulting model parameters are shown in Table II. Example of a fragment of a fisheye image and the distortion-corrected version of it is shown Fig. 10. It is notable that they contain voids caused by the truncated format of the original image. However, these voids did not interfere with further experimentation because they left a sufficient field of view unaffected.

In the experimental setup, a stereo vision system module is affixed to the table's edge, and the table is marked with 25 cm increments for positioning the target surface, which is then moved accordingly. To illustrate, when the target surface is situated 50 cm away from the camera, the expected depth is also 50 cm.

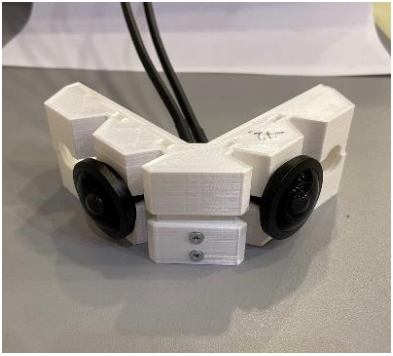

**Figure 8**: Orthogonally divergent stereo vision system prototype

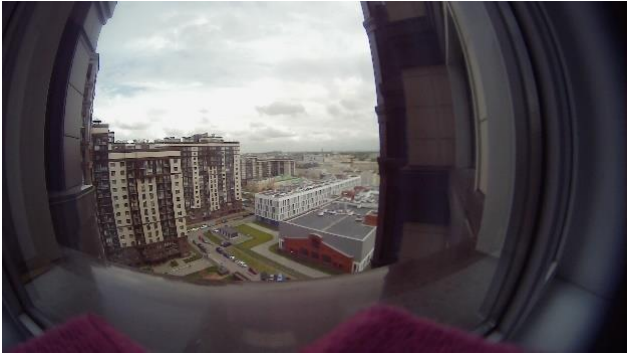

**Figure 9**: An image from one of cameras. The lens circle is not fully inscribed in the frame and is off-center, thereby reducing the effective area available for distortion removal. Vignetting around edges is also noticeable and can interfere with the search for matches. This increases the contribution of the sensor to the overall error.

The depth-quality evaluation was performed, and the results are displayed in Fig.11. The x-axis shows the depth to the plane in terms of the number of stereopair baseline lengths, and the y-axis shows the root-mean-square (RMS) error between the estimated depth and the expected depth.

We examined how the RMS error of the depth estimation varied with the depth for different distortion models: Scaramuzza, Mei, and Kannala-Brandt. We used a second order polynomial regression model to fit the data. The vertical lines show the standard deviation of the depth distribution in the pointcloud. Each line represents a different model and is calculated from the average of all the textures.

As can be seen in Fig. 11, the best depth estimation result was achieved by Kannala-Brandt model (the smallest error), while Mei and Scaramuzza models perform similarly but with lower performances compared to Kannala-Brandt model. This result demonstrates the effectiveness of the Kannala-Brandt model for our proposed method for distortion removal for divergent stereo cameras.

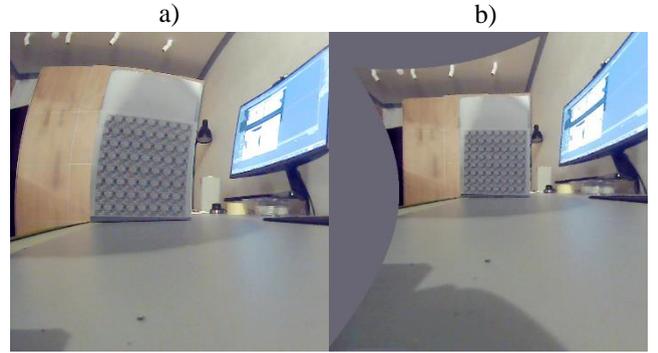

**Figure 10**: a) fragment of the original fisheye image, b) the distortion corrected version.

TABLE II: Real camera calibration results for Mei, Kannala-Brandt, and Scaramuzza models

| Model | Mei | Kannala-Brandt | Scaramuzza |
|---|---|---|---|
| Software | CamOdoCal | CamOdoCal | MATLAB |
| MRE | 0.417 | 0.418 | 0.341 |
| Parameters (left) | $\xi=2.404$, $p1=4.76*10\text{e-}4$, $p2=2.97*10\text{e-}4$, $k1=-0.115$, $k2=3.01$ | $k2=3.78*10\text{e-}4$, $k3=4.48*10\text{e-}5$, $k4=5.79*10\text{e-}3$, $k5=-2.41*10\text{e-}3$ | $a0=647$, $a1=-6.46*10\text{e-}4$, $a2=3.31*10\text{e-}7$, $a3=-3.02*10-10$ |
| Parameters (right) | $\xi=1.678$, $p1=4.54*10\text{e-}4$, $p2=-5.15*10\text{e-}4$, $k1=-0.115$, $k2=3.01$ | $k2=3.80*10\text{e-}4$, $k3=1.89*10\text{e-}3$, $k4=-1.20*10\text{e-}3$, $k5=-1.22*10\text{e-}4$ | $a0=648$, $a1=-6.46*10\text{e-}4$, $a2=3.31*10\text{e-}7$, $a3=-1.92*10\text{e-}10$ |

We conducted a simulation (virtual environment) and an experiment using a real divergent stereo camera to evaluate the RMS error of the depth estimation after applying distortion removal with Kannala-Brandt model. We compared these results with the RMS error of the depth estimation of conventional stereo cameras without camera divergence reported in [17], using the same camera parameters as our experiment.

After analyzing the results shown in Fig. 12, it was evident that the conventional scenario displayed the lowest depth error estimation, as projected. This is due to the stereo cameras being aligned with the optical axis, producing images with minimal distortion. However, the simulated case, which used divergent stereo cameras, had a lower error rate compared to the experimental setup with the same specifications. This indicates that although attempts were made to minimize factors such as camera errors, measurement conditions, and surface properties in the experimental case, they still had an impact. These factors were eliminated in the simulated scenario, resulting in lower error rates. The findings substantiate the proposed method's validity for removing distortion.

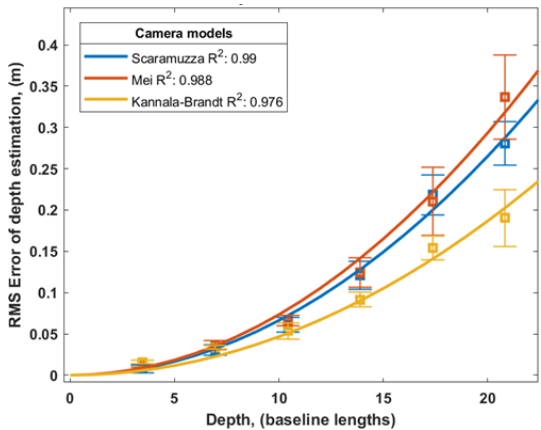

**Figure 11**: Depth quality evaluation of real orthogonally divergent stereo cameras using Mei, Kannala-Brandt and Scaramuzza models.

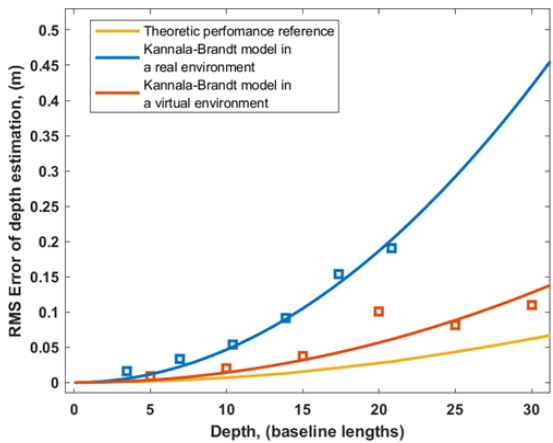

**Figure 12**: Depth quality evaluation. Theoretic performance reference refers to the performance expected under ideal conditions from a stereopair with parameters set according to the VPC stereopair in the prototype.

## V. CONCLUSION

In this paper, a method for distortion removal was proposed for orthogonally divergent fisheye cameras. The performance of the proposed method was evaluated in both simulated and experimental environments, and the results were compared with stereo cameras that have parallel optical axes. The findings indicate that the proposed method effectively removes distortion and provides accurate depth estimation. Overall, the proposed method presents a viable solution for addressing lens distortion in stereo vision systems. Further work should focus on optimizing and improving the accuracy of the distortion correction method and developing a method for automatic stereo calibration for different camera placement configurations. There is also a need to test this system for SLAM applications.

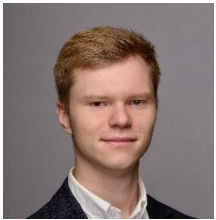
Matvei Panteleev received bachelor's degree in robotics and mechatronics from Peter the Great St. Petersburg Polytechnic University in 2022. He is a master student in autonomous systems at Aalto university. His research interests are computer vision, robotics and control systems.

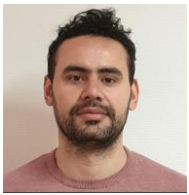
**Houari Bettahar** received his bachelor's degree in Electrical and Electronic Engineering from the Institut de Génie Électrique et Électronique (IGEE), Boumerdes, Algérie in 2011. He received his master's degree in control system and information technology in 2014 from the University of Grenoble, France and his Ph. D. degree in Automatic Control and Robotics in 2019 from the University of Bourgogne-Franche-Comté, Besançon, France. Since 2018, he has been an teaching and research Officer at University of Franche-Comté, Besançon, France working in the AS2M (Automatic Control and MicroMechatronic Systems) department of FEMTO-ST Institute. Currently, he is a postdoctoral researcher at Aalto university. His research interests are the control of microrobotic, gel- like matter manipulation, and artificial fibers fabrication using robotics and automation.